\documentclass[5p,times]{elsarticle}
\usepackage{ecrc}

\volume{00}

\firstpage{1}

\journalname{Manufacturing Letters}

\runauth{L.-W. Shih et al.}

\jid{MFGLET}

\usepackage{amsmath}

\usepackage{amssymb}

\usepackage[figuresright]{rotating}

\usepackage[bookmarks=false]{hyperref}
    \hypersetup{colorlinks,
      linkcolor=blue,
      citecolor=blue,
      urlcolor=blue}

\begin{document}
\begin{frontmatter}



\dochead{54th SME North American Manufacturing Research Conference (NAMRC 54, 2026)}%

\title{Near-Field Perception for Safety Enhancement of Autonomous Mobile Robots in Manufacturing Environments}


\author[a]{Li-Wei Shih\fnref{equal}}
\author[a]{Ruo-Syuan Mei\fnref{equal}}
\author[b]{Jesse Heidrich}
\author[b]{Hui-Ping Wang}
\author[b]{Joel Hooton}
\author[b]{Joshua Solomon}
\author[c]{Jorge Arinez}
\author[c]{Guangze Li \corref{cor1}}
\author[a]{Chenhui Shao\corref{cor1}}

\fntext[equal]{These authors contributed equally to this work.}

\address[a]{Department of Mechanical Engineering, University of Michigan, MI 48109, USA}
\address[b]{Manufacturing Technology Development, General Motors R\textup{\&}D, Warren, MI 48092, USA}
\address[c]{Materials \textup{\&} Manufacturing Systems Research, General Motors R\textup{\&}D, Warren, MI 48092, USA}

\begin{abstract}
Near-field perception is essential for the safe operation of autonomous mobile robots (AMRs) in manufacturing environments. Conventional ranging sensors such as light detection and ranging (LiDAR) and ultrasonic devices provide broad situational awareness but often fail to detect small objects near the robot base. To address this limitation, this paper presents a three-tier near-field perception framework. The first approach employs light-discontinuity detection, which projects a laser stripe across the near-field zone and identifies interruptions in the stripe to perform fast, binary cutoff sensing for obstacle presence. The second approach utilizes light-displacement measurement to estimate object height by analyzing the geometric displacement of a projected stripe in the camera image, which provides quantitative obstacle height information with minimal computational overhead. The third approach employs a computer vision-based object detection model on embedded AI hardware to classify objects, enabling semantic perception and context-aware safety decisions. All methods are implemented on a Raspberry Pi 5 system, achieving real-time performance at 25 or 50 frames per second. Experimental evaluation and comparative analysis demonstrate that the proposed hierarchy balances precision, computation, and cost, thereby providing a scalable perception solution for enabling safe operations of AMRs in manufacturing environments.
\end{abstract}

\cortext[cor1]{Corresponding authors: Guangze Li (guangze.li@gm.com), Chenhui Shao (chshao@umich.edu).}

\end{frontmatter}



\section{Introduction}
\label{sec:intro}

Autonomous mobile robots (AMRs) are now commonly used in manufacturing for material transport, parts staging, and line-side delivery in manufacturing \cite{Lackner2024,dhanda2025reviewing, wang2025multi}. They move through dynamic, human-shared spaces \cite{Villani2018,Lasota2017,li2024safe}, so perception and decision making close to the robot are central to productivity and safety. As deployments scale, plants expect AMRs to operate reliably across cluttered aisles, mixed lighting, and changing floor conditions \cite{sousa2025obstacle,bonci2021human}.

Many safety-critical interactions occur within the first half meter around the AMR chassis, where small components, dropped tools, cables, and human limbs can appear unexpectedly. In general, safety monitoring for AMRs, particularly in manufacturing environments, has been investigated only to a limited extent. A related but distinct application area is autonomous driving, where various sensing modalities, including 3D point clouds and ultrasonic waves, have been explored. For example, light detection and ranging (LiDAR)-based perception has been widely explored for autonomous navigation. A 3D obstacle-detection and tracking framework for mobile vehicles was developed in \cite{saha20243d}, showing strong performance in dynamic environments. A real-time LiDAR-camera fusion method was presented in \cite{liu2023real}, which improved detection accuracy through multimodal sensing. Ultrasonic sensing has also been investigated in related studies. For instance, the Simultaneous Calibration and Navigation (SCAN) framework proposed in \cite{gualda2019simultaneous} enabled precise localization and mapping in multi-sensor robot setups using multiple ultrasonic local positioning systems. \citet{nesti2023ultra} used multiple ultrasonic transducers mounted on a vehicle bumper for short-range object detection, though the approach was limited by coarse angular resolution and difficulty distinguishing object types.

Despite these advances, existing methods developed for autonomous driving primarily address mid- to long-range perception and do not provide sufficient precision for near-field safety, which is critical in human-robot shared environments \cite{Lackner2024}. Moreover, such approaches offer only global situational awareness and often lack the spatial resolution and low latency required to detect subtle or low-profile obstacles near the robot base. Therefore, reliable near-field perception tailored to AMRs is critically needed to prevent collisions and enable responsive motion planning in shared work environments.



To meet this need, this paper proposes a hierarchical three-tier framework consisting of three complementary approaches designed to balance speed, precision, and interpretability in near-field perception for AMRs. The first approach, a light-discontinuity detection method, enables rapid binary cutoff detection by identifying interruptions in a projected laser stripe, providing an immediate safety response when an obstacle is present. The second approach, a light-displacement-based height estimation method, estimates object height through the geometric shift of the projected stripe in the image plane. This allows the system to distinguish between small debris and larger obstacles. The third approach, a computer vision-based object-detection method enhanced by synthetic data generation for limited near-field perception data availability, introduces semantic understanding by classifying detected objects into predefined categories. With this information, the AMR can execute context-specific actions based on object type and associated risk. Experimental evaluation validated the effectiveness of all three approaches: the first achieved reliable obstacle detection across common industrial objects, the second demonstrated an average height estimation error of 17.8 mm (RMSE) for representative samples, and the third achieved 100\% mAP@0.5 accuracy in object classification. Together, these approaches form a progressive structure that evolves from geometric to semantic perception, adaptable to a range of industrial requirements in diverse manufacturing environments. 

To the best of the authors’ knowledge , this is the first study to systematically investigate near-field perception for AMRs through a set of complementary geometric and semantic sensing approaches. The developed methods are implemented on an affordable Raspberry Pi 5 platform and experimentally evaluated for detection accuracy, latency, and hardware cost. Finally, a comparative analysis identifies cost-performance trade-offs and practical considerations for deploying near-field perception in manufacturing environments.

The remainder of this paper is organized as follows. Section~\ref{sec:method} outlines the problem definition and hardware layout of the sensing system. Section~\ref{sec:method1} presents the light-cutoff detection approach. Section~\ref{sec:method2} describes the light-displacement height estimation method. Section~\ref{sec:method3} introduces the object-detection approach and its deployment on embedded hardware. Section~\ref{sec:discussion} evaluates the cost, time efficiency, and limitations, and discusses potential future extensions of the proposed three-tier approaches. Finally, Section~\ref{sec:conclusion} summarizes the key findings and concluding remarks.

\section{Method overview}
\label{sec:method}

This section introduces the overall architecture of the proposed near-field perception framework. The system integrates low-cost sensing modules around the AMR perimeter to detect nearby obstacles and classify potential hazards. The following subsections describe the hardware configuration, the predefined object-response logic, and the hierarchical sensing strategy that links all three approaches within a unified framework.

\subsection{System setup and hardware layout}
\label{sec:system-setup}
\begin{figure}[h]
    \centering
    \includegraphics[width=0.95\linewidth]{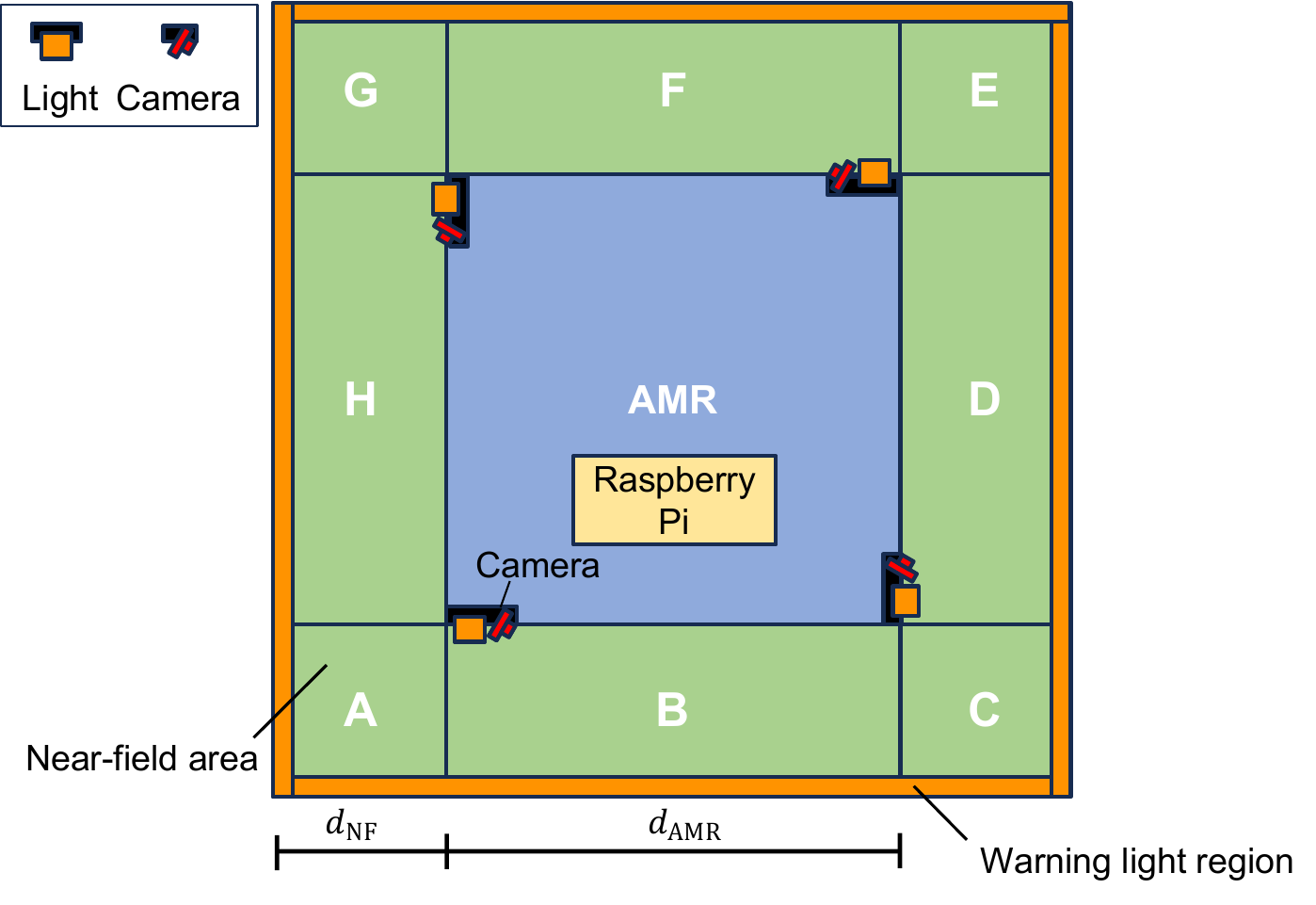}
    \caption{Top-view schematic of the AMR showing the near-field detection layout. The AMR footprint ($d_{\text{AMR}}$) is surrounded by a near-field detection range ($d_{\text{NF}}$) and a safety buffer zone illuminated by warning lights.}
    \label{fig:robot_schema}
\end{figure}

Figure \ref{fig:robot_schema} presents the top-view schematic of the hardware configuration for the near-field perception system. The AMR footprint (blue region) has a characteristic length denoted by $d_{\text{AMR}}$, while the surrounding detection zone (green region) extends outward by a distance $d_{\text{NF}}$. This zone is partitioned into eight monitoring regions, labeled A through H, distributed along the four sides of the robot perimeter. Each side is equipped with a Raspberry Pi camera module and a laser warning light, where each camera-light pair monitors two adjacent regions. For instance, the bottom-left module monitors regions B and C, while the bottom-right module monitors regions D and E. The integration of multiple camera-light module pairs minimizes blind spots and enhances system reliability, ensuring trustworthy co-existence with human operators on the factory floor.

\begin{table*}[t]
    \centering
    \caption{Action table defining AMR behavior according to object type and size}
    \label{tab:action_table}
    \vspace{6pt}
    \begin{tabular}{p{2.3cm}p{5.2cm}p{5.2cm}p{2.8cm}}
        \hline
        \textbf{Category} & \textbf{Large object ($\ge 50$\,mm)} & \textbf{Small object ($< 50$\,mm)} & \textbf{System action} \\
        \hline
        Human & Hand, foot, knee, leg, sitting or fallen operator & --- & Stop \\
        Tools & Power drill, wrench, hammer, screwdriver & Nuts, bolts, washers, markers, gloves & Continue (small); Stop or Reroute (large) \\
        Materials & Boxes, pallets, wooden planks, containers & Paper, tape, plastic film, straps & Continue (small); Stop (large) \\
        Parts & Trim panels, metal brackets, extrusions & Clips, fasteners, connectors & Stop \\
        Vehicles & Forklift, cart, other AMRs & --- & Stop or Reroute \\
        Environment & Cables, cones, pipes, signage & Chair/table legs, portable devices & Stop \\
        Safety / PPE & Barrier, cone, heavy PPE item & Safety glasses, gloves, jackets & Stop or Reroute \\
        \hline
    \end{tabular}
\end{table*}

\begin{figure*}[t]
    \centering
    \includegraphics[width=0.8\linewidth]{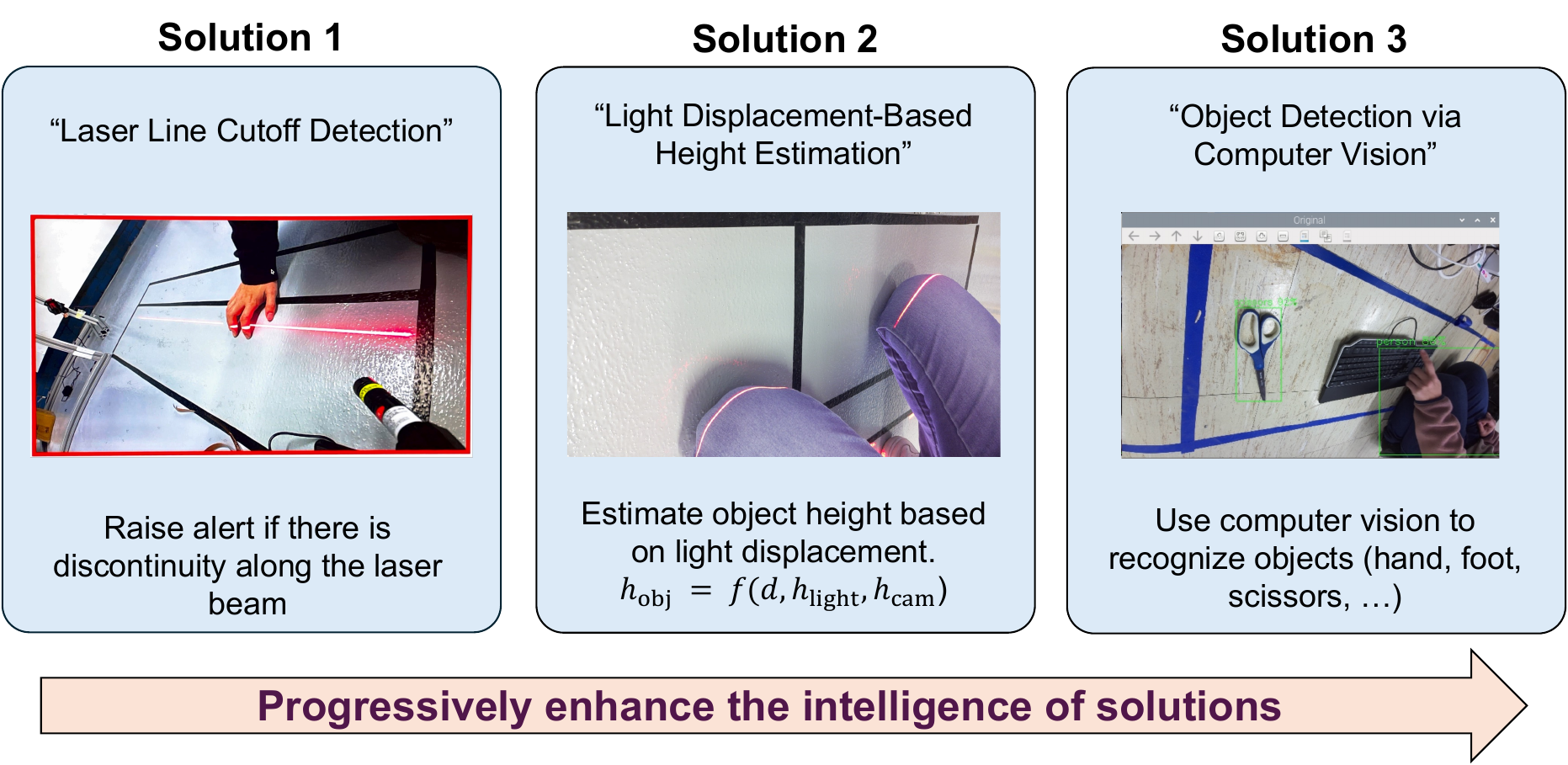}
    \caption{Overview of the three proposed near-field perception approaches.}
    \label{fig:method_overview}
\end{figure*}

\subsection{Action table}

To ensure consistent and safe behavior in dynamic manufacturing environments, a predefined action table is established to translate near-field detections into appropriate robot responses. Table~\ref{tab:action_table} summarizes the predefined response strategy for objects detected in the near-field zone. Objects are categorized into \textit{large} ($\geq50$~mm) and \textit{small} ($<50$~mm) groups across seven classes: human, tools, materials, parts, vehicles, plant infrastructure, and safety items. The AMR operates in three possible response modes:  (1) run over, (2) stop and warn, and (3) go around, depending on the object’s size, category, and potential risk to human operators or equipment. The specific response triggered depends on the level of perceptual intelligence available, allowing the AMR to react based on object presence, geometric size, or semantic category.

This action logic follows a hierarchical safety protocol: detections associated with human presence trigger immediate stops, whereas small non-hazardous debris is ignored to maintain operational efficiency. Objects classified as structural or mobile obstacles initiate a stop-and-warn response, allowing the robot to either reroute or await operator confirmation. This rule-based framework provides a balance between safety assurance and continuous navigation for AMR operations in manufacturing environments.

\subsection{Overview of the three-tier perception strategy}
\label{sec:three-tier solutions}

Using the hardware configuration described in Section \ref{sec:system-setup}, three solutions for near-field perception are proposed, as illustrated in Figure \ref{fig:method_overview} and summarized in Table~\ref{tab:approach_summary}. Approach 1, ``Laser Line Cutoff Detection,'' detects discontinuities in the laser beam projected across the detection zone and raises an alert when the beam is interrupted by an object. Approach 2, ``Light Displacement-Based Height Estimation,'' estimates object height through the relationship $h_{\text{obj}}=f(d, h_{\text{light}},h_{\text{cam}}),$ where the height is calculated from the observed light displacement $d$, laser mounting height $h_{\text{light}}$, and camera mounting height $h_{\text{cam}}$. Approach 3, ``Object Detection via Machine Vision Algorithm,'' employs computer vision algorithms to identify specific object types (e.g., hands, feet, scissors) and generates detection decisions based on object classification. These three approaches progressively enhance detection intelligence from simple presence detection to geometric estimation and ultimately to semantic object recognition. The following sections present detailed implementation and evaluation of each solution.

\begin{table}[t]
    \centering
    \caption{Overview of near-field sensing approaches, associated hardware, and key capabilities.}
    \label{tab:approach_summary}
    \vspace{6pt}
    \footnotesize
        \begin{tabular}{p{0.1cm}p{2.3cm}p{2.3cm}p{2.3cm}}
        \hline
         & \textbf{Approach} & \textbf{Hardware} & \textbf{Capability} \\
        \hline
        1 & Light cutoff detection &
        Laser diode, Raspberry~Pi~5, camera &
        Detects object presence (binary). \\
        
        2 & Light displacement &
        Same as Approach~1 &
        Estimates object height. \\
        
        3 & Object detection &
        Raspberry~Pi~5, AI~HAT+, camera &
        Recognizes object category. \\
        \hline
    \end{tabular}
\end{table}

\section{Approach 1: laser line cutoff detection}
\label{sec:method1}

Approach 1 identifies the presence of obstacles within the near-field zone through detecting interruptions in a projected light stripe. This binary detection mechanism serves as the foundation of the three-tier hierarchy by providing rapid, low-latency awareness of any object crossing the safety boundary. The method's simplicity enables real-time implementation with minimal hardware cost and computation.

\subsection{Light projection setup}
\begin{figure}[t]
    \centering
    \includegraphics[width=0.8\linewidth]{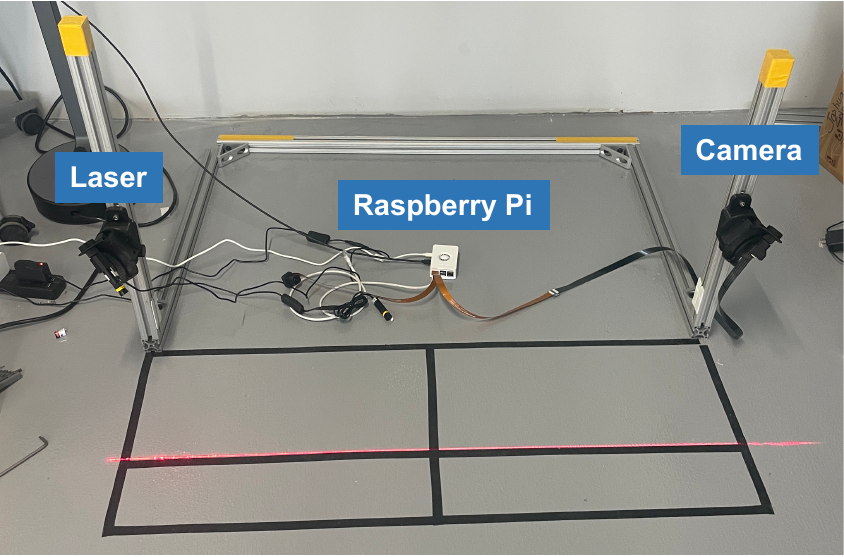}
    \caption{Hardware setup for approach 1. The laser and camera are mounted on an adjustable fixture.}
    \label{fig:sol1_hardware}
\end{figure}

A 653~nm laser diode is mounted on one side of the AMR. The laser beam has a rated power of 5 mW, which classifies it as a Class 3R device. Figure~\ref{fig:sol1_hardware} shows the laser setup. The laser is tilted downward to cover the full detection length. However, because of the limited vertical space available for installation, a centrally mounted laser cannot fully cover the required detection range. Mounting the laser on one side resolves this limitation but introduces a drawback: the far end of the projected beam is farther from the source, causing the stripe to appear blurred and wider compared with the near end. This effect requires additional compensation in the detection algorithm.

\begin{figure}[t]
    \centering
    \includegraphics[width=\linewidth]{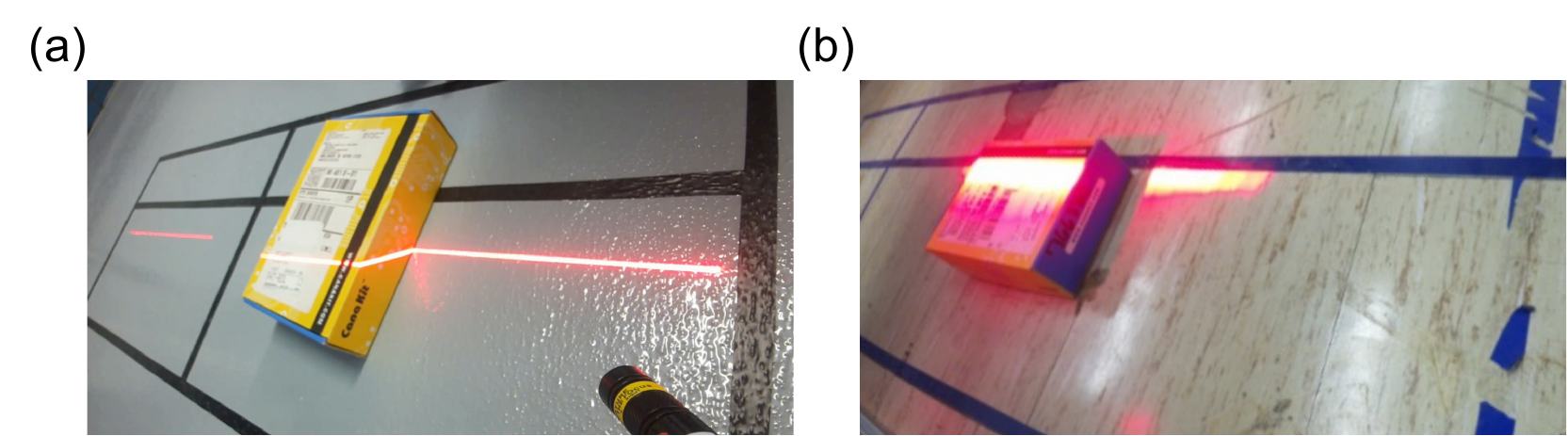}
    \caption{Comparison of illumination patterns for (a) laser-based and (b) LED-based light projection.}
    \label{fig:sol1_light}
\end{figure}

In cases where a Class 3R laser is not permitted, an alternative red-light LED can be used. Figure~\ref{fig:sol1_light} compares the laser-based and LED-based approaches. The LED projects a broader and less intense illumination pattern on the floor, which complicates segmentation. The reduced contrast between the illuminated region and the background makes it more challenging to separate the light band from the surrounding surface, particularly when the floor or nearby objects have similar color tones.

\subsection{Vision system}
\label{sec:method1_setup}

A Raspberry Pi 5 is selected as the processing unit for this system, offering a balance of low cost and high performance suitable for embedded vision tasks. The system employs a Raspberry Pi Camera Module 3 operating at a resolution of 2304~$\times$1296 pixels and a frame rate of approximately 50 frames per second. The camera is mounted on one side of the AMR at a height of approximately 50~cm from the ground and tilted downward to capture the floor area adjacent to the robot. This downward-tilted configuration allows the camera to effectively observe the near-field zones B and C, as illustrated in Figure\ref{fig:robot_schema}. The placement ensures complete coverage of the target detection region while maintaining compact integration with the robot chassis. This setup provides sufficient spatial detail for laser line analysis and discontinuity detection while achieving real-time performance on the Raspberry Pi~5 platform.

\subsection{Line discontinuity detection algorithm}
\label{sec:method1_algo}

\begin{figure}
    \centering
    \includegraphics[width=0.85\linewidth]{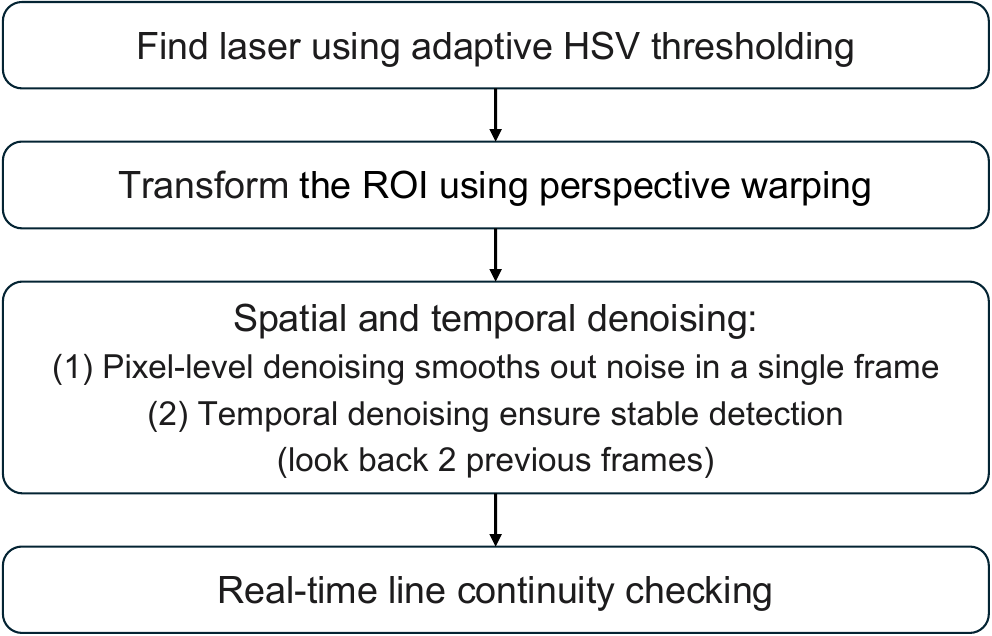}
    \caption{Processing pipeline for line cutoff detection.
}
    \label{fig:sol1_pipeline}
\end{figure}

Figure~\ref{fig:sol1_pipeline} illustrates the proposed pipeline for Approach~1. The algorithm consists of four main stages: color thresholding, region of interest (ROI) geometry warping, spatial and temporal denoising, and continuity checking. These steps convert raw camera input into a stable binary signal that represents the presence or absence of a laser line discontinuity.

\textbf{Color thresholding and line detection:}
To detect the projected laser stripe, the RGB image is first converted to the Hue-Saturation-Value (HSV) color space. The hue and saturation channels are adaptively thresholded to isolate pixels corresponding to the 653~nm laser wavelength. Morphological operations remove small spurious noise. The resulting binary mask is processed using a Hough Line Transform to identify the central trajectory of the beam and estimate its orientation in the frame.

\textbf{ROI geometry warping:}
Because the camera is mounted at an oblique angle relative to the floor, the captured image experiences perspective distortion. A homography transformation is applied to warp the region of interest into a top-down view, ensuring consistent pixel-to-millimeter scaling across the field of view. The geometric mapping parameters are obtained during calibration using a planar target with known dimensions.

\textbf{Spatial and temporal denoising:}
Spatial denoising smooths pixel intensity variations in a single frame by applying a running average kernel corresponding to a 5~mm physical window. Temporal denoising stabilizes detection results by averaging over consecutive frames. A larger temporal window increases stability but also introduces processing delay. In this work, the current frame and two previous frames are used, which provides a balanced compromise between responsiveness and robustness. At a capture rate of 50 frames per second, this corresponds to a 20~ms delay in effective response time.

\textbf{Continuity checking:}
The final step evaluates the continuity of the detected line along its longitudinal axis. A discontinuity is flagged when the maximum observed gap between line segments exceeds a predefined threshold specified by system safety requirements, typically five millimeters. When a discontinuity is detected, a trigger signal is sent to the decision module for immediate alerting and motion control.

\subsection{Results}
\label{sec:method1_result}

Approach 1 was experimentally validated using all objects listed in Table~\ref{tab:action_table}. Tests were performed with both single objects and mixed scenes, where multiple items were placed along the projected light line at a distance of approximately 50~cm from the AMR. Under standard laboratory lighting conditions, the system operated at 50~frames~per~second and generated a warning signal whenever at least one object was detected within the projected light curtain.

\begin{figure}[h]
    \centering
    \includegraphics[width=\linewidth]{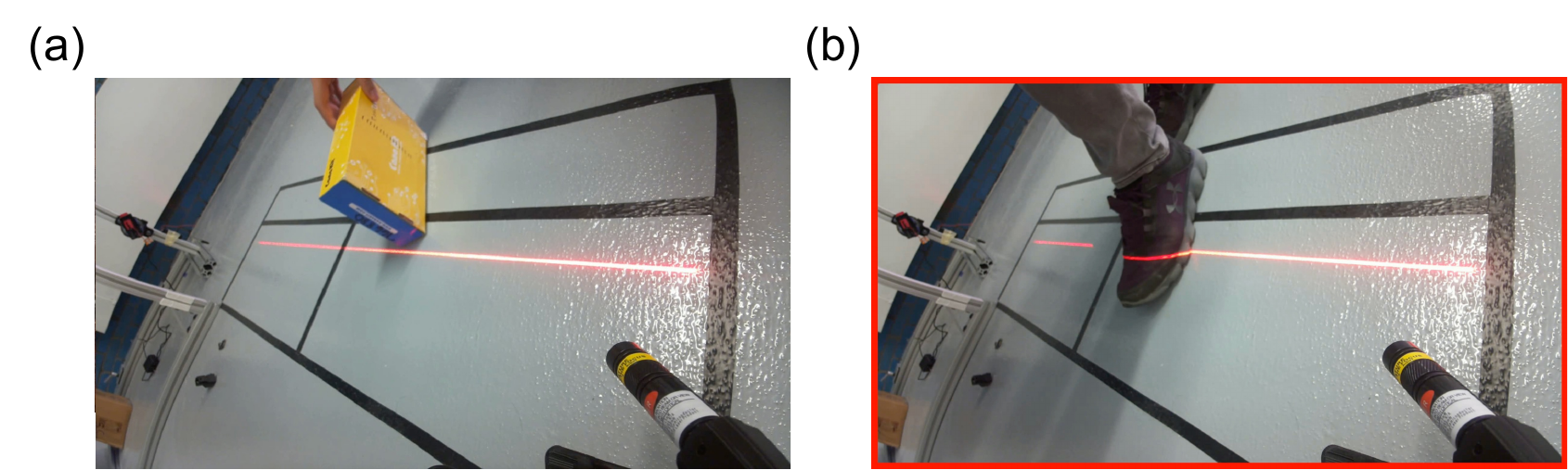}
    \caption{Example results from Approach 1, showing detection of line discontinuities when (a) no objects block the projected beam and (b) objects interrupt the projected beam.}
    \label{fig:sol1_result}
\end{figure}

All tested objects, including human body parts, tools, and plant-floor components, were evaluated at four orientations (0°, 90°, 180°, and 270°) relative to the AMR. Example detection results are shown in Figure~\ref{fig:sol1_result}. In Figure~\ref{fig:sol1_result}(a), a box is placed near the projected laser stripe without making contact; the algorithm correctly ignored the shadow and produced no false alarm. In Figure~\ref{fig:sol1_result}(b), a human shoe interrupts the laser line, resulting in a positive detection. The system successfully detected all medium and large objects across all orientations. Extremely small or thin objects, such as rubber bands and narrow zip ties, were occasionally missed, particularly near the far end of the projection line where the pixel-to-millimeter resolution decreases due to perspective distortion. These undetected cases accounted for less than five percent of the total trials and occurred primarily outside the optimal detection region.

Overall, the results demonstrate that the proposed light projection setup provides robust and repeatable performance for near-field cutoff detection. The system consistently responded to the presence of both human and non-human objects while maintaining operation at 50~FPS. The slight loss of sensitivity for sub-centimeter objects in the distant region is primarily attributed to the limited pixel-to-millimeter resolution of the camera at the far end of the projection field. Future work will explore higher-resolution imaging or multi-camera configurations to improve coverage of small and thin objects.

\section{Approach 2: light displacement-based height estimation}
\label{sec:method2}

This approach extends the discontinuity detection method introduced in Section~\ref{sec:method1}. Instead of only identifying the presence of an object that interrupts the projected light, the method estimates the object’s height by analyzing the displacement of the projected line in the camera image. This allows precise measurement of object elevation within the near-field region, although it imposes stricter requirements on the relative placement of the camera and light source.

\subsection{Height estimation}
\label{sec:method2_algo}

The concept of height estimation is derived from structured-light techniques commonly used in 3D scanning, where object height is determined from the geometric relationship between the camera and the projected light. A 653~nm laser line is projected across the floor, and when an object enters the detection region, the illuminated stripe shifts upward in the camera image due to geometric displacement. The object height can thus be expressed as a function of the observed displacement \(d\), laser mounting height \(h_\text{light}\), and camera mounting height \(h_\text{cam}\), as defined by
\vspace{-15pt}
\begin{equation}
    h_\text{obj} = f(d,\, h_\text{light},\, h_\text{cam}).
    \label{eq:hobj_overview}
\end{equation}

By measuring the pixel offset between the baseline floor reference and the observed stripe, \(h_\text{obj}\) can be estimated through a calibrated geometric mapping. In this formulation, \(d_\text{light}\) and \(d_\text{obj}\) represent the horizontal distances from the camera to the laser source and to the detected object, respectively. The parameters \(x_\text{corner}\) and \(x_\text{obj}\) correspond to the pixel coordinates of the baseline edge and object location in the image plane, while \(f_x\) denotes the focal length in pixel units obtained from camera calibration, ensuring consistency with the image-plane measurements.

\begin{figure}
    \centering
    \includegraphics[width=\linewidth]{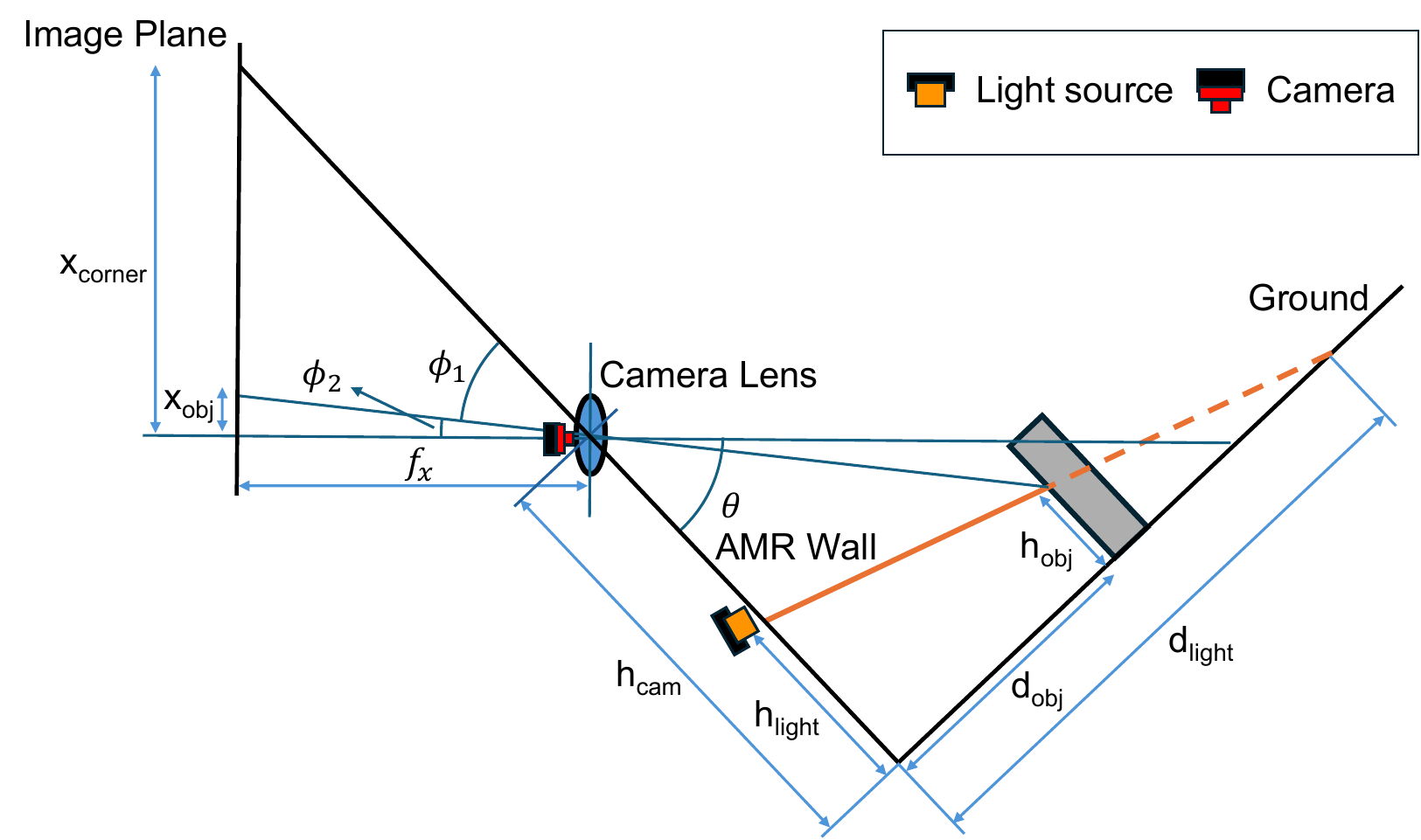}
    \caption{Geometric model of light displacement for object height estimation.}
    \label{fig:sol2_height_calc}
\end{figure}

Figure \ref{fig:sol2_height_calc} illustrates the geometric relationship among the camera, laser source, and object surface used for height estimation. The projected laser line forms a measurable displacement in the image plane when an object enters the detection region, and this shift corresponds to the object height through the known mounting geometry. Three governing equations are derived from these geometric relationships, as shown in Equations \eqref{eq:phi1}--\eqref{eq:tanphi1}.
\vspace{-15pt}
\begin{gather}
    \phi_1 = \theta - \phi_2 =
    \tan^{-1}\!\left(\frac{x_\text{corner}}{f_x}\right) -
    \tan^{-1}\!\left(\frac{x_\text{obj}}{f_x}\right),
    \label{eq:phi1} \\[3pt]
    \frac{h_\text{light}}{d_\text{light}} =
    \frac{h_\text{obj}}{d_\text{light} - d_\text{obj}},
    \label{eq:similar_triangle} \\[3pt]
    \tan \phi_1 =
    \frac{d_\text{obj}}{h_\text{cam} - h_\text{obj}}.
    \label{eq:tanphi1}
\end{gather}
By substituting Equations~\eqref{eq:similar_triangle} and~\eqref{eq:tanphi1}, a closed-form expression for object height is obtained:
\vspace{-12pt}
\begin{equation}
    h_\text{obj} = 
    h_\text{light}\frac{d_\text{light} - h_\text{cam}\tan\phi_1}
    {d_\text{light} - h_\text{light}\tan\phi_1}.
    \label{eq:hobj}
\end{equation}
Combining Equations~\eqref{eq:phi1} and~\eqref{eq:hobj} yields an analytical mapping between the observed line displacement in the image plane and the corresponding physical object height. All geometric parameters, including camera height, laser height, and projection angle, are fixed during calibration, making this formulation suitable for real-time height computation on the embedded Raspberry~Pi~5 system. The resulting \(h_\text{obj}\) values serve as input to the near-field safety system, which differentiates small debris from taller obstacles and determines whether the robot should run over or reroute around the object.

\subsection{Camera placement}
\label{sec:camera-placement}
Accurate height estimation through light displacement requires careful determination of the camera and laser light mounting positions and orientations. When the projected light stripe intersects an object, the magnitude of displacement observed in the camera image strongly depends on the camera’s placement. A larger displacement improves measurement resolution and thus enhances the precision of height estimation.

\begin{figure}[h]
    \centering
    \includegraphics[width=.85\linewidth]{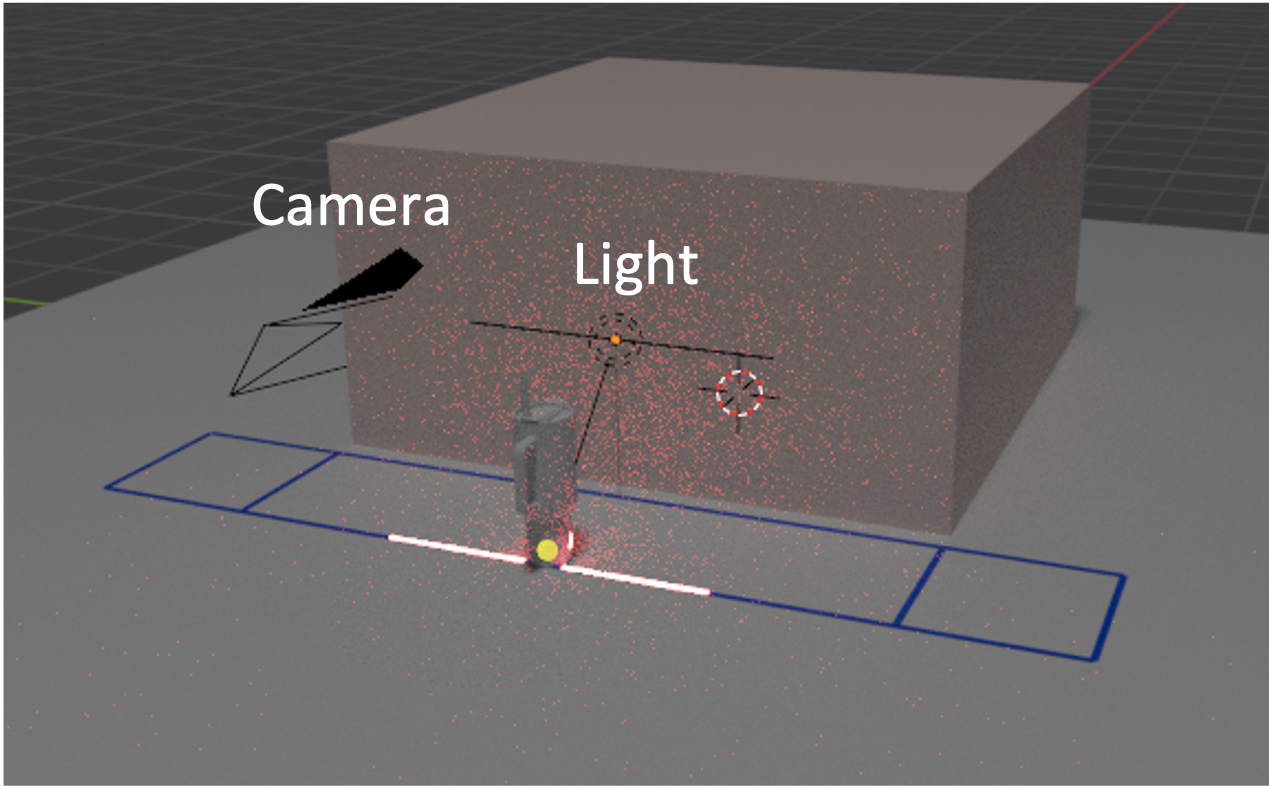}
    \caption{Digital twin in Blender used for modeling AMR geometry and testing camera–laser configurations.}
    \label{fig:blender-env}
\end{figure}

Rather than conducting time- and labor-intensive physical adjustments in the actual system, a digital twin approach is adopted using the physics-based simulation engine  \cite{blender2024} to model the AMR hardware configuration, as shown in Figure \ref{fig:blender-env}. The simulation environment replicates the AMR system using a cube geometry, with the floor plane (blue line) representing three near-field perception regions. The camera and laser light parameters, including focal length, resolution, field of view, light power, color, shape, and size, are calibrated to match the real-world hardware specifications. This digital twin enables rapid evaluation of different sensor configurations and generates synthetic images for algorithm development.

\begin{figure}[h]
    \centering
    \includegraphics[width=.85\linewidth]{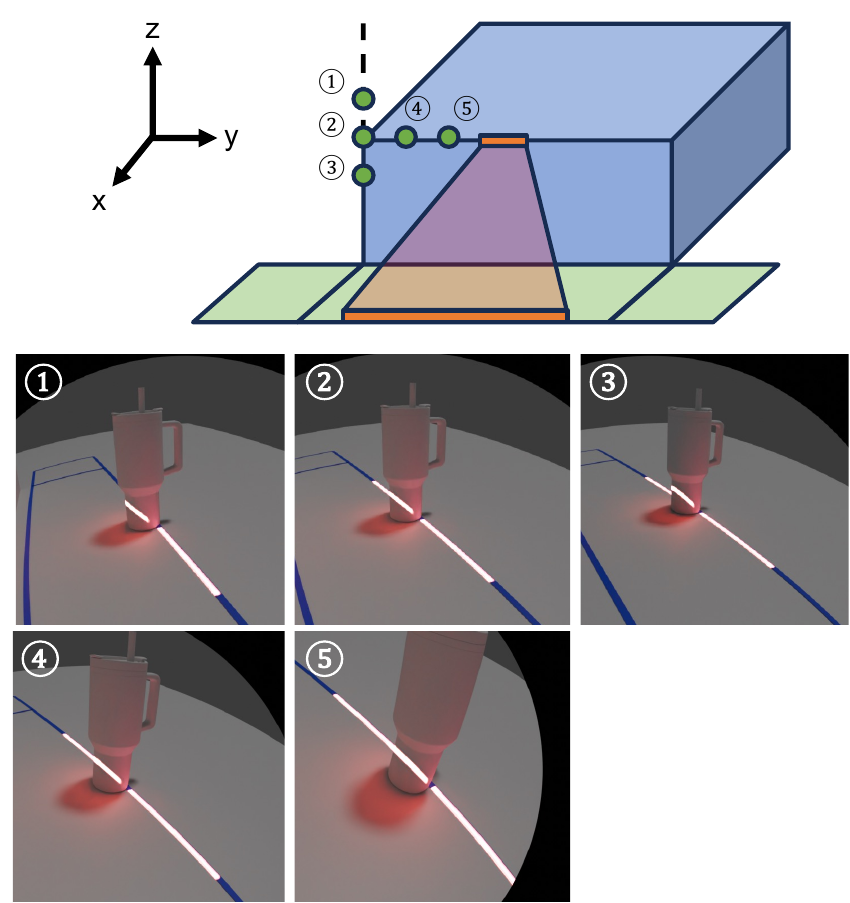}
    \caption{Evaluation of camera placement along the z-axis (positions 1–3) and y-axis (positions 2, 4, 5) relative to the AMR edge in the simulated environment.}
    \label{fig:blender-views}
\end{figure}

Figure~\ref{fig:blender-views} illustrates the rendered camera views as the camera position is varied along the y-axis and z-axis relative to the AMR edge. When the camera moves along the y-axis (positions 2, 4, and 5), the magnitude of displacement in the projected stripe remains unchanged. In contrast, shifting the camera along the z-axis (positions 1–3) produces noticeable differences: when the camera is positioned higher than the laser, the stripe shifts inward in the image, whereas when it is lower, the stripe shifts outward. The greater the vertical offset between the camera and the laser, the larger the displacement observed in the image. Consequently, to achieve higher measurement sensitivity, the camera and laser should be separated along the z-axis, while movement along the y-axis has negligible impact on displacement magnitude.

\subsection{Results}
\label{sec:method2_result}

The hardware setup for Approach~2 is identical to that used in Approach~1, as described in Section~\ref{sec:method1_result}. Specifically, the camera is mounted at a height of 40~cm and the laser at 20~cm above the floor to create a distinct light stripe across the near-field detection zone. To measure each object's height, the object was positioned such that the projected light intersected its top edge, defining the measurement point. Using the geometric formulation in Equation~\eqref{eq:hobj}, the object height was derived from the measured pixel displacement of the projected stripe in the captured image.

The height estimation performance is summarized in Table~\ref{tab:height_result}. Across seven representative samples, the system achieved a root-mean-square error (RMSE) of 17.8~mm. Most deviations occurred for reflected objects such as boxes and aluminum extrusions, where the laser stripe appeared diffuse and introduced greater pixel-level uncertainty. For less diffuse surfaces, such as the knee and hand, the estimation accuracy improved to within 10~mm. Overall, the achieved accuracy is sufficient to discriminate object height categories within the near-field region, confirming that the light displacement approach effectively complements the discontinuity-based detection method. Future work will focus on three dimensional model that can accommodate tilted camera angle and image processing to improving estimation on reflective objects to reduce error from height estimation.

\begin{table}[t]
    \centering
    \small
    \caption{Height estimation results for light displacement-based approach}
    \vspace{6pt}
    \label{tab:height_result}
    \begin{tabular}{lccc}
        \hline
        \textbf{Object} & \textbf{Ground Truth} & \textbf{Predicted} & \textbf{Error} \\
        & \textbf{(mm)} & \textbf{(mm)} & \textbf{(mm)} \\
        \hline
        Box (1) & 40 & 63 & +23 \\
        Box (2) & 60 & 78 & +18 \\
        Box (3) & 80 & 104 & +24 \\
        Aluminum extrusion & 40 & 59 & +19 \\
        Hand & 50 & 60 & +10 \\
        Shoes & 60 & 79 & +19 \\
        Knee & 120 & 113 & $-$7 \\
        \hline
        \textbf{RMSE} & & & \textbf{17.8} \\
        \hline
    \end{tabular}
\end{table}

\section{Approach 3: object recognition via machine vision algorithm}
\label{sec:method3}

The third approach introduces semantic understanding into the near-field perception. Instead of relying on geometric cues alone, a machine learning model is trained to recognize and classify factory-floor objects directly from camera images. This enables context-aware decision-making aligned with the predefined action table, allowing the AMR to distinguish between tools, materials, and human limbs.

\subsection{Synthetic dataset generation}
\label{sec:SDG}
\begin{figure}[h]
    \centering
    \includegraphics[width=\linewidth]{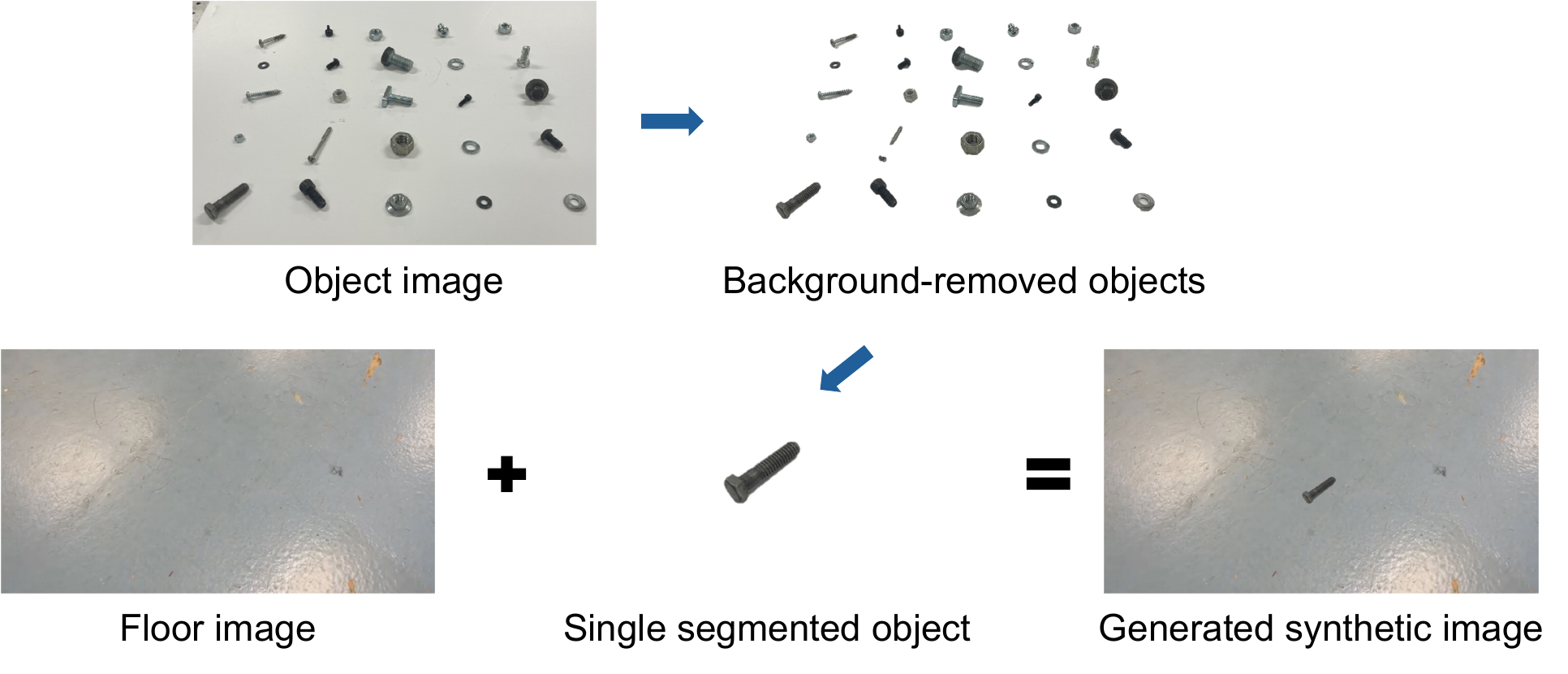}
    \caption{Synthetic dataset generation pipeline for industrial environment objects.}
    \label{fig:sdg}
\end{figure}

\begin{figure}[h]
    \centering
    \includegraphics[width=0.85\linewidth]{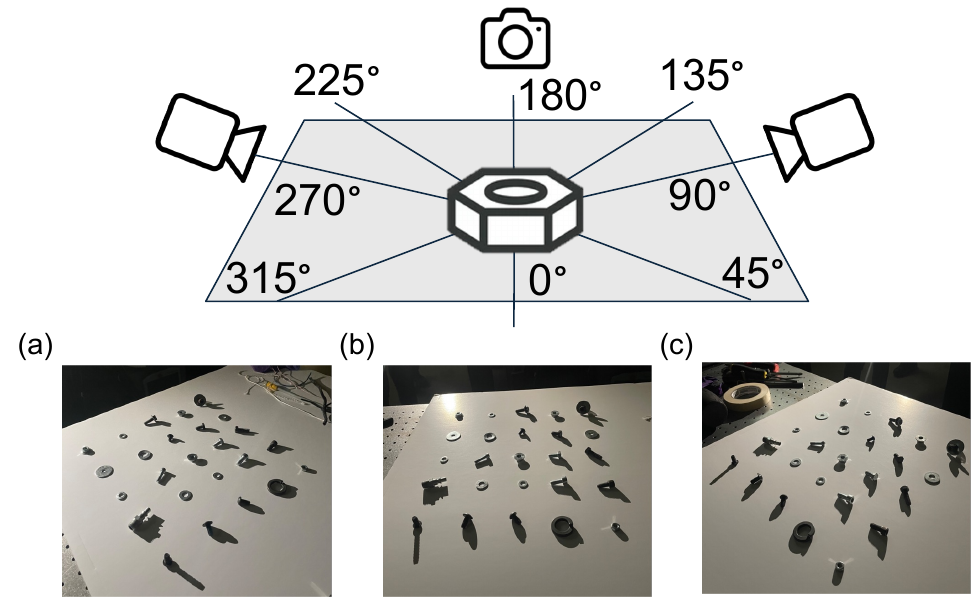}
    \caption{Example of multi-angle data acquisition from eight orientations for each object category (a) $45^\circ$, (b) $90^\circ$, (c) $135^\circ$.}
    \label{fig:sdg-angle}
\end{figure}

\begin{figure}[h]
    \centering
    \includegraphics[width=0.85\linewidth]{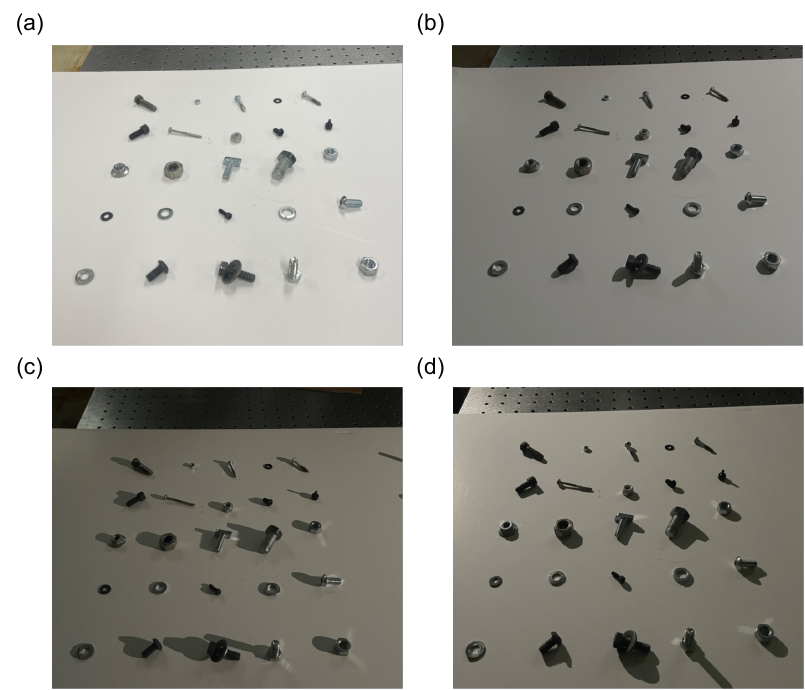}
    \caption{Example of lighting variation in the dataset to enhance robustness of model training (a) regular lighting (b) reduced lighting (c) lighting at $225^\circ$ (d) $45^\circ$.}
    \label{fig:sdg-light}
\end{figure}

To the best of the authors’ knowledge, no publicly available dataset exists for near-field object detection in manufacturing environments. As such, images of common factory floor objects were collected and overlaid onto floor plane images captured by the AMR camera to generate a synthetic dataset for model training. Synthetic data generation has increasingly been used in smart manufacturing to overcome shortages of labeled training data and to improve the robustness of computer-vision models under diverse operating conditions \cite{mei2025synthetic}. The synthetic data generation pipeline is illustrated in Figure \ref{fig:sdg}. To improve robustness against variations in object pose and illumination, data were acquired from eight viewing angles at 45° intervals (0°–315°) for each object class (Figure \ref{fig:sdg-angle}) and under five distinct lighting conditions (Figure \ref{fig:sdg-light}). Each object image was then segmented and background-removed, producing approximately 2,800 clean object samples across 70 unique items. In parallel, 2,000 floor images were randomly cropped from AMR video recordings to serve as backgrounds. Synthetic scenes were generated by randomly varying object number, position, and orientation on these floor images, yielding a final dataset of 5,000 synthetic images for model training and validation. This multi-angle and multi-illumination data collection strategy ensures that the trained detection model can reliably recognize objects regardless of their orientation or lighting condition within the near-field detection zone.

\subsection{Object-detection model design and implementation}
\label{sec:object_detection}

The object-detection model is implemented using the YOLOv5m architecture for real-time semantic recognition of near-field objects. Among the 5,000 synthetic image, 3,500 are used for training, 500 for validation and 1,000 are used for testing. The model is initialized from pretrained yolov5m.pt weights and fine-tuned for 50 epochs with a batch size of 16. After training, the model achieves a mean average precision (mAP@0.5) of 100 percent on the validation set, demonstrating strong generalization across all ten object categories.

For deployment, the trained network is compiled to the HALIO 8L format and executed on a Raspberry Pi 5 equipped with a Raspberry Pi AI HAT+. The inference pipeline includes image capture and resizing, tensor packing for HALIO 8L, hardware-accelerated inference, and CPU-side post-processing. The system operates at a sustained rate of approximately 25 frames per second. This configuration allows immediate object recognition and responsive decision output according to action table in Table \ref{tab:action_table}. The object-detection approach enables the system to distinguish small objects from human hands or shoes of similar height, supporting more context-aware decision making in manufacturing environments.

\subsection{Results}
\label{sec:method3_result}

The trained YOLOv5m model was evaluated on the synthetic dataset and verified using real images captured by the AMR-mounted camera. Quantitative results are summarized in Table~\ref{tab:obj_metrics_full}. The detector achieved a mean average precision (mAP@0.5) of 100\% and mAP@0.5:0.95 of 99\% across ten object categories. Precision and recall values for all classes exceeded 0.85, indicating reliable identification of both small and large parts commonly present on the factory floor. On-device inference using the Raspberry Pi AI HAT+ achieved an average latency of 40 ms per frame, corresponding to a sustained input rate of 25 frames per second.

\begin{figure}
    \centering
    \includegraphics[width=0.85\linewidth]{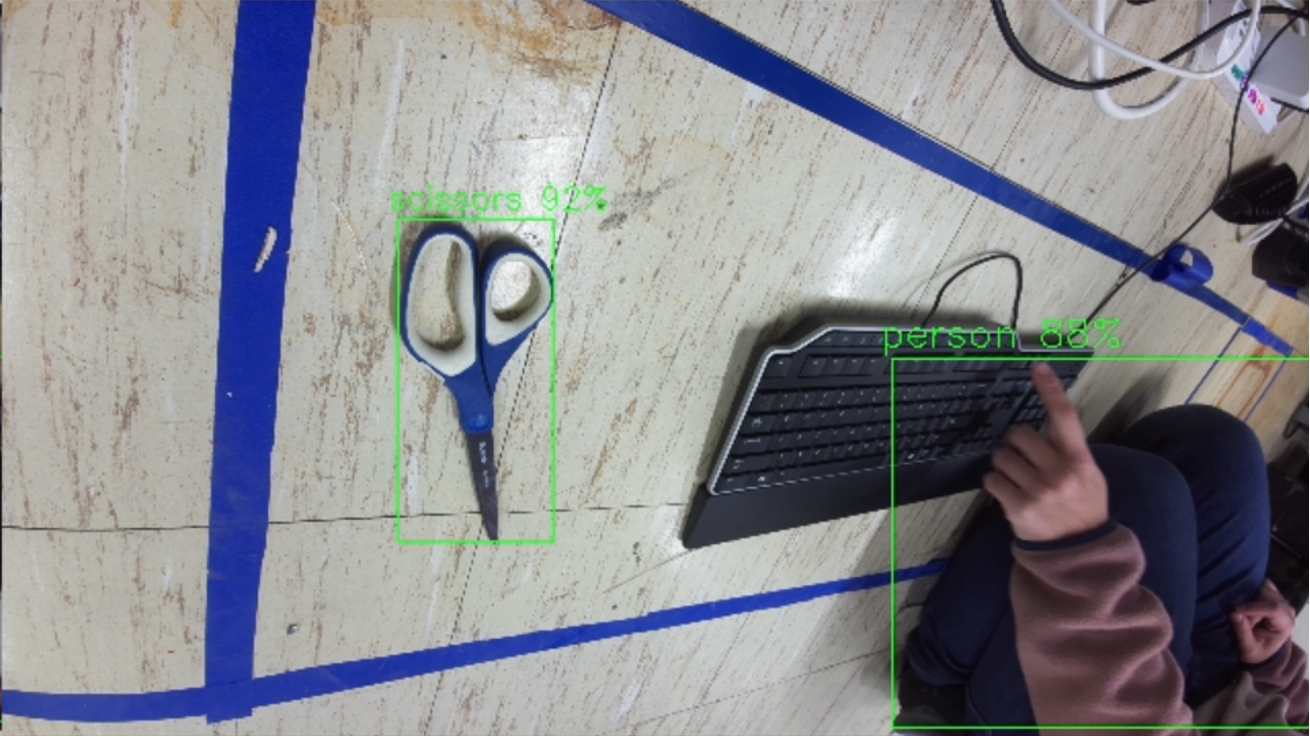}
    \caption{Example inference results of the YOLOv5m model showing accurate detection of a human and tool within the near-field region.}
    \label{fig:sol3_object_detection}
\end{figure}

Representative qualitative results are shown in Figure~\ref{fig:sol3_object_detection}, which illustrates accurate detection of a human foot and a metallic component within the near-field region. The model correctly localized and classified both objects with high confidence, demonstrating its ability to distinguish human body parts from surrounding industrial materials. These results confirm that the object-detection module provides consistent and real-time semantic awareness, supporting the decision-making layer for near-field safety monitoring.

\begin{table}[h]
    \centering
    \small
    \caption{Per-class performance of YOLOv5m object detection model}
    \vspace{6pt}
    \label{tab:obj_metrics_full}
    \begin{tabular}{lccc}
        \hline
        \textbf{Class} & \textbf{Precision} & \textbf{Recall} & \textbf{F1 Score} \\
        \hline
        Bolt & 0.98 & 0.92 & 0.95 \\
        Nut & 0.97 & 0.90 & 0.94 \\
        Washer & 0.96 & 0.88 & 0.92 \\
        Connector & 0.95 & 0.89 & 0.92 \\
        Wire & 0.94 & 0.87 & 0.90 \\
        Screwdriver & 0.93 & 0.85 & 0.89 \\
        Wrench & 0.94 & 0.87 & 0.90 \\
        Marker & 0.91 & 0.82 & 0.86 \\
        Zip tie & 0.92 & 0.83 & 0.87 \\
        Allen key & 0.95 & 0.86 & 0.90 \\
        \hline
        \textbf{mAP@0.5} & \multicolumn{3}{c}{1.00} \\
        \hline
    \end{tabular}
\end{table}

\section{Discussion}
\label{sec:discussion}

This section discusses the comparative characteristics of the three approaches in terms of system cost, computational efficiency, and application suitability. The analysis highlights how sensing complexity increases alongside functional capability, offering practical guidelines for selecting approaches in manufacturing environments.

\begin{table*}[t]
    \centering
    \small
    \caption{Comparison of cost, performance, and capabilities across three near-field perception approaches}
    \label{tab:discussion_comparison}
    \vspace{6pt}
    \begin{tabular}{p{1.5cm}p{1.5cm}p{1.5cm}p{1.2cm}p{3.5cm}p{6.5cm}}
        \hline
        \textbf{Approach} & \textbf{Setup time (min)} & \textbf{Frame rate (FPS)} & \textbf{Latency (ms)} & \textbf{Capability} & \textbf{Main limitations} \\
        \hline
        1 & 20 & 50 & 40 & Detects presence only (binary cutoff) & Cannot classify or size objects; thin objects at far end may be missed; subject to Class 3R safety policy \\
        2 & 60 & 50 & 45 & Estimates object height & Sensitive to reflective surfaces and far-end distortion; cannot distinguish human parts from small tools \\
        3 & 10 & 25 & 40 & Full semantic classification and action-table execution & Lower FPS; limited to trained classes; affected by low lighting \\
        \hline
    \end{tabular}
\end{table*}

\subsection{Cost and system complexity}
\label{sec:cost}

The cost and complexity of the system scale with the sensing capability of each approach. Table~\ref{tab:discussion_comparison} compares the three near-field perception approaches in terms of cost, performance, and capabilities. Approach~1 uses a low-cost Class~3R laser diode, a camera module, and a Raspberry~Pi~5, totaling approximately \$207, making it the most economical configuration. Approach~2 employs identical hardware but introduces geometric alignment requirements between the laser and camera for light-displacement calibration, maintaining the same cost but slightly higher structural complexity. Approach~3 removes the laser entirely and instead deploys the YOLOv5m model using a Raspberry~Pi~5 with a HALIO~AI~HAT+ hardware acceleration and the same camera module, totaling approximately \$257. While its hardware cost is moderately higher, this configuration achieves the most advanced sensing capability by adding semantic interpretation and the ability to execute the complete action table described in Table~\ref{tab:action_table}. The increase in overall cost and system complexity is therefore directly correlated with the intelligence and flexibility of the system.

\subsection{Time efficiency and computational performance}
\label{sec:performance}

The computational performance and system delay is important to real-time warning system. The performance results in Table~\ref{tab:discussion_comparison} show that all three approaches maintain real-time operation on the Raspberry~Pi~5 platform. Approaches~1 and~2 operate at 50~frames~per~second with latencies of 40~ms and 45~ms, respectively, allowing immediate cutoff or displacement analysis without perceptible delay. The minor latency increase in Approach~2 arises from the pixel-shift computation required to calculate height displacement. Approach~3, accelerated by the HALIO~AI~HAT+, processes frames at 25~frames~per~second with an average latency of 40~ms. Setup effort follows a similar hierarchy: Approach~1 required approximately twenty minutes for mechanical alignment and threshold adjustment, Approach~2 required about one hour for camera and light height calibration, and Approach~3 was operational within ten minutes once the trained model was uploaded. These results demonstrate that the AI-based approach, while computationally heavier, offers the shortest setup and most scalable deployment process.

\subsection{Applicability and limitations}
\label{sec:limit}

Table~\ref{tab:discussion_comparison} compares the cost, performance, and capabilities of the three proposed perception methods. Each method has specific operational constraints that define its ideal application domain. Approach~1 is suitable for low-cost binary cutoff detection, where the primary goal is to determine whether any object interrupts the light curtain. However, it cannot classify or size objects and therefore cannot follow the predefined action table. Approach 2 extends geometric sensing capability by introducing height estimation based on light displacement. This enhancement allows partial differentiation between small debris and larger obstacles, providing additional geometric context to near-field perception. The experimental results achieve a 17.8~mm RMSE in height prediction. The accuracy can be further improved through refined image processing, particularly to enhance estimation performance on reflective objects. Despite this improvement, it remains unable to reliably distinguish between a small tool and a human hand of similar height and suffers from reduced accuracy on reflective surfaces. Approach~3 overcomes these limitations by providing full semantic understanding of the scene, achieving 100~percent mAP@0.5 accuracy and mapping each detection to the appropriate response category in the action table. The trade-offs include sensitivity to lighting conditions and dependence on class coverage within the trained dataset.

In summary, these three-tier approaches form a sequence of increasing cost, intelligence, and adaptability: from binary cutoff detection to geometry-aware estimation and finally to full semantic perception. Manufacturers can select the most suitable solution according to safety, performance, and budget requirements.


\subsection{Future work}
\label{sec:future_work}

Each approach presents opportunities for improvement and further study. 
For Approach~1, a straightforward improvement is to install an additional camera on the opposite side of the AMR to capture the laser line from two perspectives. This dual-view configuration would compensate for perspective distortion near the far end of the projection, reduce the pixel-to-millimeter resolution loss, and improve the detection of thin or low-profile objects. Combining the two views could also allow limited triangulation to estimate object width without increasing computational cost.

For Approach~2, future work will focus on enhancing the reliability of height estimation through improved calibration and multi-wavelength projection \cite{chang2024real,lyu2024multi}. Employing structured patterns rather than a single line could provide greater spatial redundancy and reduce sensitivity to surface reflectivity.

Approach~3 offers the greatest potential for system-level expansion. 
Future research will extend the object-detection framework to a multimodal perception architecture that combines visual detection with geometric and temporal cues from Approaches~1 and~2. This fusion would allow the system to retain the speed and simplicity of laser-based sensing while leveraging semantic context for robust decision-making under variable lighting and partial occlusion. Additional directions include deploying lightweight Transformer-based backbones to improve energy efficiency \cite{li2022exploring}, expanding domain adaptation to enable rapid model transfer across different plant environments \cite{khodabandeh2019robust,fatima2024advancing}, and exploring continuous learning on the AMR for adaptive perception in dynamic manufacturing settings \cite{shmelkov2017incremental,wang2021wanderlust}.

Beyond the current hardware setup, a promising direction is the integration of high-precision 3D scanners with excellent near-field sensing capabilities. Although such scanners are widely used in manufacturing quality inspection \cite{yang2021data,biehler2023plural,mei2024deep}, their application to near-field safety enhancement for AMRs remains limited. The fine-scale 3D information has the potential to support more accurate decision-making, though associated computational challenges may need to be addressed \cite{engelcke2017vote3deep}. Moreover, while multi-modal sensing fusion has shown effectiveness in a range of recent applications, including safety-critical autonomous systems \cite{chen2022multimodal}, contextual object detection \cite{zang2025contextual}, and real-time multi-modal detection \cite{sharma2020yolors}, its potential for improving near-field safety perception in AMRs is still underexplored. Future work could investigate complementary sensing modalities, such as 3D point clouds, thermal, infrared, and ranging sensors, to enhance near-field collision avoidance and safety performance across diverse manufacturing environments.


\section{Conclusion}
\label{sec:conclusion}

This study presented a three-tier framework for near-field perception on AMRs, spanning geometric and semantic perception within a scalable architecture. The first approach provided a simple and binary solution for real-time laser line cutoff detection. The second approach extended capability by estimating object height from laser-stripe shifts, adding geometric context with minimal computational overhead. The third method further advanced perception by introducing semantic understanding through embedded AI hardware, supporting category-based decision-making aligned with a predefined action table.

Experimental results confirmed real-time performance across the three methods on a Raspberry~Pi~5 platform, with frame rates ranging from 25 to 50~FPS. Together, these methods demonstrate a scalable sensing hierarchy for AMRs that balances precision, cost, and interpretability. The proposed framework offers a practical foundation for intelligent near-field perception to enable safe operations of AMRs in manufacturing environments.

\section*{Acknowledgements}
The authors gratefully acknowledge the financial support of General Motors (GM) and the technical assistance provided by Zhongyou Zhao and Weitian Zhou from GM during the experimental phase of this study.





\bibliographystyle{elsarticle-num-names} 
\bibliography{references}

@article{dhanda2025reviewing,
  title={Reviewing human-robot collaboration in manufacturing: Opportunities and challenges in the context of industry 5.0},
  author={Dhanda, Mandeep and Rogers, Benedict Alexander and Hall, Stephanie and Dekoninck, Elies and Dhokia, Vimal},
  journal={Robotics and Computer-Integrated Manufacturing},
  volume={93},
  pages={102937},
  year={2025},
  publisher={Elsevier}
}

@article{wang2025multi,
  title={Multi-robot collaborative manufacturing driven by digital twins: advancements, challenges, and future directions},
  author={Wang, Gang and Zhang, Cheng and Liu, Sichao and Zhao, Yongxuan and Zhang, Yingfeng and Wang, Lihui},
  journal={Journal of Manufacturing Systems},
  volume={82},
  pages={333--361},
  year={2025},
  publisher={Elsevier}
}

@article{li2024safe,
  title={Safe human--robot collaboration for industrial settings: a survey},
  author={Li, Weidong and Hu, Yudie and Zhou, Yong and Pham, Duc Truong},
  journal={Journal of Intelligent Manufacturing},
  volume={35},
  number={5},
  pages={2235--2261},
  year={2024},
  publisher={Springer}
}

@article{Lackner2024,
  author    = {Lackner, M. and Wurhofer, D. and Tscheligi, M.},
  title     = {Review of automated mobile robots in intralogistics},
  journal   = {Journal of Industrial Information Integration},
  year      = {2024},
  publisher = {Elsevier}
}

@article{Villani2018,
  author    = {Villani, V. and Pini, F. and Leali, F. and Secchi, C.},
  title     = {Survey on human--robot collaboration in industrial settings},
  journal   = {Mechatronics},
  volume    = {61},
  pages     = {1--13},
  year      = {2018}
}

@article{Lasota2017,
  author    = {Lasota, P. A. and Fong, T. and Shah, J. A.},
  title     = {A survey of methods for safe human--robot interaction},
  journal   = {Foundations and Trends in Robotics},
  volume    = {5},
  number    = {4},
  pages     = {261--349},
  year      = {2017},
  publisher = {Now Publishers}
}

@article{sousa2025obstacle,
  title={Obstacle Avoidance Technique for Mobile Robots at Autonomous Human-Robot Collaborative Warehouse Environments},
  author={Sousa, Lucas C and Silva, Yago MR and Schettino, Vin{\'\i}cius B and Santos, Tatiana MB and Zachi, Alessandro RL and Gouv{\^e}a, Josiel A and Pinto, Milena F},
  journal={Sensors},
  volume={25},
  number={8},
  pages={2387},
  year={2025},
  publisher={MDPI}
}

@article{bonci2021human,
  title={Human-robot perception in industrial environments: A survey},
  author={Bonci, Andrea and Cen Cheng, Pangcheng David and Indri, Marina and Nabissi, Giacomo and Sibona, Fiorella},
  journal={Sensors},
  volume={21},
  number={5},
  pages={1571},
  year={2021},
  publisher={MDPI}
}

@inproceedings{nesti2023ultra,
  title={Ultra-sonic sensor based object detection for autonomous vehicles},
  author={Nesti, Tommaso and Boddana, Santhosh and Yaman, Burhaneddin},
  booktitle={Proceedings of the IEEE/CVF conference on computer vision and pattern recognition},
  pages={210--218},
  year={2023}
}

@article{saha20243d,
  title={3D LiDAR-based obstacle detection and tracking for autonomous navigation in dynamic environments},
  author={Saha, Arindam and Dhara, Bibhas Chandra},
  journal={International Journal of Intelligent Robotics and Applications},
  volume={8},
  number={1},
  pages={39--60},
  year={2024},
  publisher={Springer}
}

@article{gualda2019simultaneous,
  title={Simultaneous calibration and navigation (SCAN) of multiple ultrasonic local positioning systems},
  author={Gualda, David and Ure{\~n}a, Jes{\'u}s and Garcia, Juan C and Garcia, Enrique and Alcala, Jose},
  journal={Information Fusion},
  volume={45},
  pages={53--65},
  year={2019},
  publisher={Elsevier}
}

@article{liu2023real,
  title={Real time object detection using LiDAR and camera fusion for autonomous driving},
  author={Liu, Haibin and Wu, Chao and Wang, Huanjie},
  journal={Scientific Reports},
  volume={13},
  number={1},
  pages={8056},
  year={2023},
  publisher={Nature Publishing Group UK London}
}

@article{chang2024real,
  title={Real-time height measurement with a line-structured-light based imaging system},
  author={Chang, Hui and Li, Deyu and Zhang, Xiangyu and Cui, Xingchen and Fu, Zhichao and Chen, Xinyu and Song, Yongxin},
  journal={Sensors and Actuators A: Physical},
  volume={368},
  pages={115164},
  year={2024},
  publisher={Elsevier}
}

@article{lyu2024multi,
  title={Multi-wavelength structured light based on metasurfaces for 3D imaging},
  author={Lyu, Baiying and Chen, Chen and Wang, Jian and Li, Chang and Zhang, Wei and Feng, Yuxiang and Dong, Fei and Zhang, BaoShun and Zeng, Zhongming and Wang, Yiqun and others},
  journal={Nanophotonics},
  volume={13},
  number={4},
  pages={477--485},
  year={2024},
  publisher={De Gruyter}
}

@inproceedings{li2022exploring,
  title={Exploring plain vision transformer backbones for object detection},
  author={Li, Yanghao and Mao, Hanzi and Girshick, Ross and He, Kaiming},
  booktitle={European conference on computer vision},
  pages={280--296},
  year={2022},
  organization={Springer}
}

@inproceedings{khodabandeh2019robust,
  title={A robust learning approach to domain adaptive object detection},
  author={Khodabandeh, Mehran and Vahdat, Arash and Ranjbar, Mani and Macready, William G},
  booktitle={Proceedings of the IEEE/CVF international conference on computer vision},
  pages={480--490},
  year={2019}
}

@inproceedings{fatima2024advancing,
  title={Advancing industrial object detection through domain adaptation: A solution for industry 5.0},
  author={Fatima, Zainab and Zardari, Shehnila and Tanveer, Muhammad Hassan},
  booktitle={Actuators},
  volume={13},
  number={12},
  pages={513},
  year={2024},
  organization={MDPI}
}

@article{mei2025synthetic,
  title   = {Synthetic Data Generation in Smart Manufacturing Applications: A Systematic Review},
  author  = {Mei, Ruo-Syuan and Li, Guangze and Jia, Sixian and Huang, Lu and Shih, Li-Wei and Arinez, Jorge and Abell, Jeffery and Shao, Chenhui},
  journal = {SSRN Electronic Journal, preprint. Available at SSRN: http://dx.doi.org/10.2139/ssrn.5726804},
  year    = {2025},
  month   = oct,
}

@inproceedings{shmelkov2017incremental,
  title={Incremental learning of object detectors without catastrophic forgetting},
  author={Shmelkov, Konstantin and Schmid, Cordelia and Alahari, Karteek},
  booktitle={Proceedings of the IEEE international conference on computer vision},
  pages={3400--3409},
  year={2017}
}

@inproceedings{wang2021wanderlust,
  title={Wanderlust: Online continual object detection in the real world},
  author={Wang, Jianren and Wang, Xin and Shang-Guan, Yue and Gupta, Abhinav},
  booktitle={Proceedings of the IEEE/CVF international conference on computer vision},
  pages={10829--10838},
  year={2021}
}

@software{blender2024,
  title = {Blender – a 3D modelling and rendering package},
  author = {{Blender Foundation}},
  version = {4.5},
  year = {2024},
  howpublished = {\url{https://www.blender.org}},
}

@article{mei2024deep,
title = {Deep learning of 3D point clouds for detecting geometric defects in gears},
journal = {Manufacturing Letters},
volume = {41},
pages = {1324-1333},
year = {2024},
note = {52nd SME North American Manufacturing Research Conference (NAMRC 52)},
issn = {2213-8463},
author = {Ruo-Syuan Mei and Christopher H. Conway and Miles V. Bimrose and William P. King and Chenhui Shao},
}

@inproceedings{chen2022multimodal,
  title={Multimodal object detection via probabilistic ensembling},
  author={Chen, Yi-Ting and Shi, Jinghao and Ye, Zelin and Mertz, Christoph and Ramanan, Deva and Kong, Shu},
  booktitle={European Conference on Computer Vision},
  pages={139--158},
  year={2022},
  organization={Springer}
}

@article{zang2025contextual,
  title={Contextual object detection with multimodal large language models},
  author={Zang, Yuhang and Li, Wei and Han, Jun and Zhou, Kaiyang and Loy, Chen Change},
  journal={International Journal of Computer Vision},
  volume={133},
  number={2},
  pages={825--843},
  year={2025},
  publisher={Springer}
}

@article{sharma2020yolors,
  title={YOLOrs: Object detection in multimodal remote sensing imagery},
  author={Sharma, Manish and Dhanaraj, Mayur and Karnam, Srivallabha and Chachlakis, Dimitris G and Ptucha, Raymond and Markopoulos, Panos P and Saber, Eli},
  journal={IEEE Journal of Selected Topics in Applied Earth Observations and Remote Sensing},
  volume={14},
  pages={1497--1508},
  year={2020},
  publisher={IEEE}
}

@article{biehler2023plural,
  title={PLURAL: 3D point cloud transfer learning via contrastive learning with augmentations},
  author={Biehler, Michael and Sun, Yiqi and Kode, Shriyanshu and Li, Jing and Shi, Jianjun},
  journal={IEEE Transactions on Automation Science and Engineering},
  volume={21},
  number={4},
  pages={7550--7561},
  year={2023},
  publisher={IEEE}
}

@article{yang2021data,
  title={Data-driven intelligent 3D surface measurement in smart manufacturing: Review and outlook},
  author={Yang, Yuhang and Dong, Zhiqiao and Meng, Yuquan and Shao, Chenhui},
  journal={Machines},
  volume={9},
  number={1},
  pages={13},
  year={2021},
  publisher={MDPI}
}

@inproceedings{engelcke2017vote3deep,
  title={Vote3deep: Fast object detection in 3d point clouds using efficient convolutional neural networks},
  author={Engelcke, Martin and Rao, Dushyant and Wang, Dominic Zeng and Tong, Chi Hay and Posner, Ingmar},
  booktitle={2017 IEEE International Conference on Robotics and Automation (ICRA)},
  pages={1355--1361},
  year={2017},
  organization={IEEE}
}

\clearpage\onecolumn

\end{document}